# Loopy Belief Propagation and Gibbs Measures

**Sekhar C. Tatikonda**
Electrical Engineering and Computer Science
University of California, Berkeley
Berkeley, CA, 94720
tatikond@eecs.berkeley.edu

**Michael I. Jordan**
Computer Science and Statistics
University of California, Berkeley
Berkeley, CA 94720
jordan@cs.berkeley.edu

## Abstract

We address the question of convergence in the loopy belief propagation (LBP) algorithm. Specifically, we relate convergence of LBP to the existence of a weak limit for a sequence of Gibbs measures defined on the LBP's associated computation tree. Using tools from the theory of Gibbs measures we develop easily testable sufficient conditions for convergence. The failure of convergence of LBP implies the existence of multiple phases for the associated Gibbs specification. These results give new insight into the mechanics of the algorithm.

## 1 Introduction

The loopy belief propagation (LBP) algorithm is an algorithm developed for computing approximate marginal statistics over graphs with cycles. This algorithm has had notable success, especially for iterative channel decoding of turbo codes and low density parity check codes. However, the behavior of LBP is poorly understood. In particular, it is not always known if this algorithm will converge.

Many new methods that generalize the basic algorithm have been developed. Some of these methods include Kikuchi based methods [6], the tree reparameterization technique [4], and the double-loop scheme [7]. Also it has been shown that the LBP algorithm can be viewed an iterative descent down an associated Bethe free energy. [6]

These advances notwithstanding, a complete understanding of the convergence properties of the original LBP algorithm is still lacking. This paper presents a new framework for analyzing the LBP algorithm. In particular we use tools developed in the study of Gibbs measures to analyze the question of convergence in the LBP algorithm. This analysis gives new insight into the mechanics of LBP.

Our analysis relies on the computation tree. This tree represents an unwrapping of the original graph with respect to the LBP algorithm. [5] The initializing messages can be represented by potentials placed at the leaves of the computation tree. We can then construct a sequence of Gibbs measures defined on the infinite computation tree. If this sequence converges then LBP converges. Our contributions are as follows:

- First, we relate LBP convergence to the existence of a weak limit for the sequence of measures defined on the corresponding computation tree.

- Second, we relate the set of all LBP fixed points to the set of all Markov chains defined on the corresponding computation tree.

- Third, we show that LBP always converges in the case when there is a unique Gibbs measure defined on the computation tree. (Conversely if LBP fails to converge then there exist multiple phases on the computation tree.)

- Finally, we provide an easily testable sufficient condition to insure convergence of LBP. In particular if

$$\max_{s \in S} \sum_{A \ni s} (|A| - 1)\delta(\Phi_A) < 2$$

then LBP will converge. (Where $\delta(\Phi)$ is a measure of the strength of the potential $\Phi$ and the sum $\sum_{A \ni s}$ is over all the neighbors of node $s$. The notation is defined in the sequel.)

## 2 Background and the LBP Algorithm

In this section we review finite Gibbs measures, the LBP algorithm, and the associated computation tree.

## 2.1 Finite Gibbs Measures

Let $S$ be a finite set of nodes. Associated with each node $i \in S$ there is a measure space $(\mathcal{X}_i, \mathcal{F}_i)$. We assume that all the $\mathcal{X}_i$ are finite.

Let
$$(\Omega, \mathcal{F}) \triangleq (\prod_{i \in S} \mathcal{X}_i, \prod_{i \in S} \mathcal{F}_i)$$

equal the product measure space. On the measure space $(\Omega, \mathcal{F})$ define an independent reference measure $\lambda = \prod_{i \in S} \lambda_i$ where each $\lambda_i$ is the uniform measure on $(\mathcal{X}_i, \mathcal{F}_i)$.

Let
$$\mathbb{S} \triangleq \{\Lambda \subset S, \Lambda \neq \emptyset\}$$

be the set of all nonempty subsets of $S$. For $A \in \mathbb{S}$ let $\Omega_A$ and $\mathcal{F}_A$ equal the restriction of $\Omega$ and $\mathcal{F}$ to $A$ respectively. Similarly $\omega_A$ represents the projection of $\omega \in \Omega$ to the set $\Omega_A$.

We will now define a Gibbs measure on $(\Omega, \mathcal{F})$. To that end we first define a potential.

**Definition 2.1** *A* potential *is a family* $\Phi = \{\Phi_A\}_{A \in \mathbb{S}}$ *of functions* $\Phi_A : \Omega \to \mathbb{R}$ *such that*

*(1) For each $A \in \mathbb{S}$ we have $\Phi_A$ is $\mathcal{F}_A$-measurable.*

*(2) For all $\Lambda \in \mathbb{S}$ and $\omega \in \Omega$ the energy*

$$H_\Lambda^\Phi(\omega) \triangleq \sum_{A \in \mathbb{S},\ A \cap \Lambda \neq \emptyset} \Phi_A(\omega) \quad \text{exists.}$$

**Definition 2.2** *The* finite Gibbs measure *is defined as*

$$\mu^\Phi(\omega) = \frac{e^{-H_S^\Phi(\omega)} \lambda(\omega)}{Z_S^\Phi}$$

*where*

$$Z_S^\Phi = \sum_{w \in \Omega} e^{-H_S^\Phi(\omega)} \lambda(\omega)$$

where $Z_S^\Phi$ is called the *partition function*.

**Definition 2.3** *The* Markov graph *associated with a Gibbs measure with potential $\Phi$ is an undirected graph $(S^\Phi, E^\Phi)$ where*

*(1) the vertices of the graph, $S^\Phi$, are the nodes in $S$*

*(2) $\{i,j\} \in E^\Phi$ if there exists a nonzero potential $\Phi_A$ with $\{i,j\} \in A$ (there are no self-loops: if $\{i,j\} \in E^\Phi$ then $i \neq j$.)*

We define the set of directed edges associated with the edge set $E^\Phi$ to be

$$\vec{E}^\Phi \triangleq \{(i,j),\ (j,i)\ :\ \{i,j\} \in E^\Phi\}.$$

It will be our convention that $\{i,j\}$ represents the *undirected* edge between node $i$ and node $j$ whereas $(i,j)$ represents the *directed* edge from node $i$ to node $j$.

A *pairwise potential* is a potential $\Phi = \{\Phi_A\}_{A \in \mathbb{S}}$ such that if $|A| > 2$ then $\Phi_A = 0$. We limit our discussion to Gibbs measures based on potentials consisting of pairwise potentials. It is shown in [5] that without loss of generality any Gibbs Measure can be represented as a Gibbs measure with pairwise potentials. This new representation, though, may lead to a large state expansion.

In many inference problems one often distinguishes between two kinds of nodes: hidden and observed. Furthermore each observed nodes is assume to be independent of all the other nodes conditioned on a particular hidden node. In this paper it will be assumed that the effects of the observations have been captured by the self-potentials: $\Phi_{\{i\}}$, $i \in S$.

## 2.2 Loopy Belief Propagation Algorithm

We now review the loopy belief propagation algorithm. Recall we are interested in computing the marginal distribution at each node of a finite Gibbs measure defined by a pairwise potential. The LBP algorithm attempts to do this be transmitting *messages* between the nodes and computing *beliefs* at each node.

We can think of the messages and beliefs as probability measures. Specifically, the *message* for the directed edge $(i,j) \in \vec{E}^\Phi$ at time $n$ is a measure :

$$m_{(i,j)}^n : \mathcal{X}_j \to [0,1]$$

such that $\sum_{w_j \in \mathcal{X}_j} m_{(i,j)}^n(\omega_j) = 1$. Similarly, the *belief* at node $i \in S$ at time $n$ is a measure:

$$b_i^n : \mathcal{X}_i \to [0,1]$$

such that $\sum_{w_i \in \mathcal{X}_i} b_i^n(\omega_i) = 1$.

We define a generic operation $\eta$ that takes a bounded, nonnegative function $f$ on a finite domain $\mathcal{X}$ and outputs its *normalization*. Specifically $\eta : f \mapsto \sum_{x \in \mathcal{X}} f(x)$.

Given a graph $(S, E)$ and any set $A \in \mathbb{S}$ let

$$\partial A \triangleq \{j \in S \setminus A\ :\ \{i,j\} \in E \text{ for some } i \in A\}$$

be the *boundary* of the set $A$ in the graph. We will abuse notation and use $\partial i$ and $\partial i \setminus j$ to represent $\partial\{i\}$ and $\partial\{i\} \setminus \{j\}$ respectively.

**Definition 2.4** *The* loopy belief propagation algorithm *consists of the following iteration on messages.*

For each $(i,j) \in \vec{E}^\Phi$:

$$m_{(i,j)}^{n+1}(\omega_j) \triangleq \eta \sum_{\omega_i \in \mathcal{X}_i} e^{-\left(\Phi_{\{i,j\}}(\omega_i,\omega_j) + \Phi_{\{i\}}(\omega_i)\right)} \prod_{k \in \partial i \setminus j} m_{(k,i)}^n(\omega_i).$$

The beliefs at time $n$ are

$$b_i^n(\omega_i) \triangleq \eta \, e^{-\Phi_{\{i\}}(\omega_i)} \prod_{k \in \partial i} m_{(k,i)}^n(\omega_i)$$

The messages are initialized to $\{m_{(i,j)}^0(\omega_j)\}_{(i,j) \in \vec{E}^\Phi}$.

**Definition 2.5** *The LBP algorithm is said to converge if there exists a unique set of messages $\{m_{(i,j)}^*\}_{(i,j) \in \vec{E}^\Phi}$ such that for each $(i,j) \in \vec{E}^\Phi$ the sequence of messages*

$$\lim_{n \to \infty} \|m_{(i,j)}^n - m_{(i,j)}^*\|_{TV} = 0.$$

*Where $\|\cdot\|_{TV}$ is the total variation norm.*

If the messages converge then clearly the beliefs converge.

**LBP on Finite Trees**

For potentials with Markov graphs that are trees LBP (more rightfully called BP in this case) converges to the true marginals. We state this result and give a representation of the measure $\mu^\Phi$ in terms of the messages.

Recall a *tree* is a singly connected, undirected graph without any loops. We sometimes single out one node $s \in S$ to be called the *root*. On the tree there is a natural distance measure $d : S \times S \to \mathbb{R}^+$, where $d(i,j)$ is the number of edges on the unique path from node $i$ to node $j$. Let $L_n^s \subset S$ be the set of nodes that are exactly a distance $n$ away from the root $s$. We will just write $L_n$ when the root $s$ is obvious.

The following result is standard and can be found in, for example, [2].

**Proposition 2.1** *Let $\Phi$ be a pairwise potential whose Markov graph $(S^\Phi, E^\Phi)$ is a tree. Then*

*(1) for any set of initial messages $\{m_{(i,j)}^0\}_{(i,j) \vec{E}^\Phi}$ the LBP algorithm converges to a unique set of messages $\{m_{(i,j)}^*\}_{(i,j) \vec{E}^\Phi}$*

*(2) for any connected subset $A \subset S^\Phi$ one has*

$$\mu^\Phi(\omega_A) = \eta \prod_{B \subseteq A} e^{-\Phi_B(\omega_A)} \prod_{i \in \partial A} m_{(i,i_A)}^*(\omega_{i_A}) \quad (1)$$

*where $i_A$ be the unique neighbor of node $i \in \partial A$ in the set $A$. Hence the beliefs converge to the true marginals.*

Note that equation (1) states that the marginal on any connected set of nodes in a tree can be determined by the potentials defined on the set and the messages transmitted across the set's boundary. A belief is just a marginal on one node.

### 2.3 The Computation Tree

We now show that $n$ iterations of the LBP algorithm on a given finite, pairwise potential, Gibbs measure can be represented as an exact BP algorithm on an associated Gibbs measure defined on a tree, specifically the *computation tree*. [5]

**Definition 2.6** *Given a pairwise potential $\Phi$ and its graph $(S^\Phi, E^\Phi)$, the associated computation tree of depth $n$ with root $s \in S^\Phi$, denoted $(\tilde{S}_n^{\Phi,s}, \tilde{E}_n^{\Phi,s})$, is defined as the tree that consists of all length $n$ paths in the graph, $(S^\Phi, E^\Phi)$, starting at $i_0 = s$, $(i_0, i_1, i_2, ..., i_n)$, that never backtrack. Specifically the tree consists of all length $n$ paths where $\{i_k, i_{k+1}\} \in E^\Phi$ and $i_k \neq i_{k+2}$.*

Figure one shows an example of a computation tree of depth three starting at node $a$. In the figure we have labeled each node in the computation tree with the associated node in the original graph.

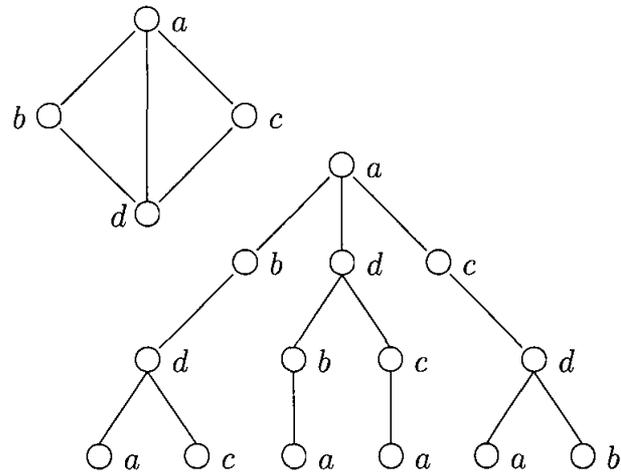

Figure 1: Example of a Computation Tree

We want to construct a Gibbs measure whose Markov graph is the computation tree. For each edge and non-leaf node we will place a potential corresponding to the potential on the original graph. We need only define the set of *boundary* self-potentials on the leaves in terms of the initializing messages. For each $(i,j) \in \vec{E}^\Phi$ let

$$\Phi_{\{i\}}^{i \to j}(\omega_i) = -\sum_{k \in \partial i \setminus j} \ln m_{(k,i)}^0(\omega_i).$$

Note that in the case when the initial messages are all set to the ones vector we have $\Phi_{\{i\}}^{i \to j} \equiv 0$.

**Definition 2.7** *Given $\Phi$, its associated computation tree, $(\tilde{S}_n^{\Phi,s}, \tilde{E}_n^{\Phi,s})$, and a set of boundary self-potentials $\{\Phi_i^{i \to j}\}_{(i,j) \in \bar{E}^\Phi}$ define the associated potential for the computation tree of depth $n$, denoted $\tilde{\Phi}^{n,s}$, as follows:*

(1) Let the map $\Gamma^n : \tilde{S}_n^{\Phi,s} \to S^\Phi$ be the map that takes node $\tilde{i}$ to its associated node $i$ in the original graph. (As constructed in definition 2.6.)

(2) For each $\{\tilde{i}, \tilde{j}\} \in \tilde{E}_n^{\Phi,s}$ let $\tilde{\Phi}^{n,s}_{\{\tilde{i},\tilde{j}\}} = \Phi_{\{\Gamma^n(\tilde{i}), \Gamma^n(\tilde{j})\}}$

(3) For each $\tilde{i} \in \tilde{S}_n^{\Phi,s}$ let

$$\tilde{\Phi}^{n,s}_{\{\tilde{i}\}} = \begin{cases} \Phi_{\{\Gamma^n(\tilde{i})\}} & \text{if } \tilde{i} \in \tilde{S}_n^{\Phi,s} \setminus L_n^s \\ \Phi_{\{\Gamma^n(\tilde{i})\}} + \Phi_{\{\Gamma^n(\tilde{i})\}}^{\Gamma^n(\tilde{i}) \to \Gamma^n(\tilde{j})} & \text{if } \tilde{i} \in L_n^s \text{ and} \\ & \text{where } \tilde{j} \text{ is the unique parent of } \tilde{i} \end{cases}$$

Let $\mu^{\tilde{\Phi}^{n,s}}(\tilde{\omega})$ be the measured determined by $\tilde{\Phi}^{n,s}$. By construction, running LBP (in this case BP) to compute the belief at the root node, $s$, of the computation tree of depth $n$ is equivalent to computing the belief at node $s$ on the original graph after $n$ iterations of LBP initialized appropriately.

## 3 Convergence of LBP and the Weak Limit

A sequence of measures $\{\mu^{\tilde{\Phi}^{n,s}}\}_{n \geq 1}$ has a *weak limit* if for each $N$ there exists a measure $\mu_N$ defined on $\tilde{S}_N^{\Phi,s}$ such that for all events $A \in \mathcal{F}_{\tilde{S}_N^{\Phi,s}}$ we have

$$\lim_{n \geq N, n \to \infty} \mu^{\tilde{\Phi}^{n,s}}(A) = \mu_N(A).$$

**Proposition 3.1** *The LBP algorithm converges if and only if the sequence of measures $\{\mu^{\tilde{\Phi}^{n,s}}\}_{n \geq 1}$ has a weak limit.*

**Proof:** We sketch the proof. Full details can be found in [3]. Using proposition 2.1 and the definition of the computation tree we have $\forall n > N$ and $\forall \zeta \in \Omega_{\tilde{S}_N^{\Phi,s}}$:

$$\mu^{\tilde{\Phi}^{n,s}}(\zeta) = \eta \prod_{B \subseteq \tilde{S}_N^{\Phi,s}} e^{-\Phi_B(\zeta)} \prod_{\tilde{i} \in \partial \tilde{S}_N^{\Phi,s}} m^{n-N}_{(\Lambda^N(\tilde{i}), \Lambda^N(\tilde{i}_{\tilde{S}_N^{\Phi,s}}))}(\zeta)$$

For fixed $N$ the boundary $|\partial \tilde{S}_N^{\Phi,s}|$ is finite. If LBP converges then the second factor above converges, as $n \to \infty$, to $\prod_{\tilde{i} \in \partial \tilde{S}_N^{\Phi,s}} m^*_{(\Lambda^N(\tilde{i}), \Lambda^N(\tilde{i}_{\tilde{S}_N^{\Phi,s}}))}(\zeta)$ and hence the weak limit exists. A similar argument proves the other direction. □

Our goal for the rest of the paper is to understand the convergence properties of the sequence of measures $\{\mu^{\tilde{\Phi}^{n,s}}\}_{n \geq 1}$. However, instead of working with a sequence of finite trees of increasing depth we will find it easier to study one infinite tree and the measures defined on it.

## 4 Gibbs Measures Over a Countable Set of Sites

Constructing a Gibbs measure over a countable set of nodes can be a tricky business. For example, there can be many Gibbs measures consistent with a given local specification provided by the potentials. This section is devoted to the construction of these Gibbs measures.

### 4.1 Specifications

We now let the set of nodes $S$ be countably infinite and redefine

$$\mathbb{S} = \{\Lambda \subset S : 0 < |\Lambda| < \infty\}$$

to be the set of all nonempty, *finite*, subsets of $S$. At every node $i \in S$ there is a finite measure space $(\mathcal{X}_i, \mathcal{F}_i)$. We construct, in the usual way, the product measure space $(\Omega, \mathcal{F}) = (\prod_{i \in S} \mathcal{X}_i, \prod_{i \in S} \mathcal{F}_i)$. We also extend the uniform reference measure $\lambda = \prod_{i \in S} \lambda_i$ on $(\Omega, \mathcal{F})$. As before we restrict ourselves to pairwise potentials. We assume that the number of neighbors at any node is finite and hence for any $\Lambda \in \mathbb{S}$ the energy $H_\Lambda^\Phi(\omega) = \sum_{A \in \mathbb{S}, A \cap \Lambda \neq \emptyset} \Phi_A(\omega)$ exists.

Because there are a countably infinite number of nodes we cannot compute the partition function by summing over all nodes. But we can discuss the partition function when conditioned on a particular boundary. Define the partition function in $\Lambda \in \mathbb{S}$ for the potential $\Phi$, boundary $\omega_{S \setminus \Lambda}$, and reference measure $\lambda$ to be

$$Z_\Lambda^\Phi(\omega) \triangleq \sum_{\zeta \in \Omega_\Lambda} e^{-H_\Lambda^\Phi(\zeta, \omega_{S \setminus \Lambda})} \lambda_\Lambda(\zeta).$$

**Definition 4.1** *Given a potential $\Phi$, $\omega \in \Omega$, and $\Lambda \in \mathbb{S}$. Then the measure*

$$\gamma_\Lambda^\Phi(\cdot \mid \omega) \triangleq \frac{e^{-H_\Lambda^\Phi(\cdot, \omega_{S \setminus \Lambda})} \lambda_\Lambda(\cdot)}{Z_\Lambda^\Phi(\omega)}$$

*is called the Gibbs distribution in $\Lambda$ with boundary $\omega_{S \setminus \Lambda}$ and potential $\Phi$. Furthermore $\gamma^\Phi = \{\gamma_\Lambda\}_{\Lambda \in \mathbb{S}}$ is called the Gibbsian specification for $\Phi$.*

Given a Gibbsian specification, which is a local description of a measure, we can ask how many measures are consistent with it. Define the set of Gibbs measures for the potential $\Phi$ to be

$$G(\gamma^\Phi) \triangleq \{\mu \in \mathcal{P}(\Omega, \mathcal{F}) : \mu(A \mid \mathcal{F}_{\Lambda^c}) = \gamma_\Lambda^\Phi(A \mid \cdot) \\ \mu - a.s. \ \forall A \in \mathcal{F} \text{ and } \Lambda \in \mathbb{S}\}.$$

The following proposition gives us an implicit characterization of the elements in $G(\gamma^\Phi)$.

**Proposition 4.1** $\mu \in G(\gamma^\Phi)$ if and only if $\mu\gamma_\Lambda^\Phi = \mu$, $\forall \Lambda \in \mathbb{S}$

**Proof:** See [1]. □

The equations $\mu\gamma_\Lambda^\Phi = \mu$, $\forall \Lambda \in \mathbb{S}$ are called the *DLR equations* after Dobrushin, Langford, and Ruelle. They state that

$$\mu\gamma_\Lambda^\Phi(\cdot \mid \omega) = \sum_\omega \mu(\omega)\gamma_\Lambda^\Phi(\cdot \mid \omega) = \mu(\cdot).$$

This is a restatement of the fact that the iterated conditional expectation equals the unconditional expectation. The DLR equations will play a very important role in our subsequent analysis. One can think of $G(\gamma^\Phi)$ as the set of all measures locally consistent with the specified potential $\Phi$. Said another way $G(\gamma^\Phi)$ is the set of all measures preserved under a countable number of proper probability kernels: $\{\gamma_\Lambda^\Phi(\cdot \mid \omega)\}_{\Lambda \in \mathbb{S}}$.

For the pairwise potential, $\Phi$, one can show that each $\mu \in G(\gamma^\Phi)$ is a *Markov field*. Specifically

$$\mu(\omega_\Lambda \mid \mathcal{F}_{\Lambda^c}) = \mu(\omega_\Lambda \mid \mathcal{F}_{\partial\Lambda}) \; \mu-a.s. \; \forall \omega_\Lambda \in \Omega_\Lambda \; \forall \Lambda \in \mathbb{S}.$$

This is sometimes called the *local Markov property*. We will discuss the *global Markov property* when we discuss Markov chains on trees in the next subsection.

**Characterization of $G(\gamma^\Phi)$**

The set $G(\gamma^\Phi)$ can either be empty, contain one measure, or contain an infinite number of measures. A potential is said to exhibit a *phase transition* if the set $|G(\gamma^\Phi)| > 1$. Phase transitions are a remarkable phenomena. For a given local specification we can have very different global behaviors. For a proof of the following proposition see [1].

**Proposition 4.2** *Let $\Phi$ be an admissible, pairwise potential defined on the countably infinite set of nodes $S$. There exists at least one Gibbs measure for an admissible pairwise potential $\Phi$.*

We have just shown that $G(\gamma^\Phi)$ is nonempty. We now discuss its structure. Clearly it is a convex set because the convex combination of two elements in $G(\gamma^\Phi)$ will certainly be a member of $G(\gamma^\Phi)$. To prove this it is enough to show that the convex combination satisfies the DLR equations.

Furthermore we can characterize the extreme points of the convex set $G(\gamma^\Phi)$. Recall an element of a convex set is *extreme* if it cannot be represented as the convex combination of other elements in the set. Each extremal element of $G(\gamma^\Phi)$ is called a *phase*. Define the *tail* sigma field to be $\mathcal{T} \triangleq \cap_{\Lambda \in \mathbb{S}} \mathcal{F}_{\Lambda^c}$.

**Proposition 4.3** *Let $\mu \in G(\gamma^\Phi)$ then the following are equivalent*

*(1) $\mu$ is extreme*

*(2) $\mu$ is trivial on the tail sigma field $\mathcal{T}$*

*(3) for all cylinder events $A$*

$$\lim_{\Lambda \in \mathbb{S}} \sup_{B \in \mathcal{F}_{\Lambda^c}} |\mu(A \cap B) - \mu(A)\mu(B)| = 0$$

**Proof:** See propositions 7.7 and 7.9 of [1]. □

An extremal measure is a mixing measure in the sense of point (3). If there is a unique measure $G(\gamma^\Phi) = \{\mu\}$ then that $\mu$ is necessarily extremal, tail-trivial, and mixing.

### 4.2 Markov Fields on Trees: Boundary Laws, Markov Chains, and Limits

We have discussed Gibbs measures defined on countable sets of nodes. We now restrict our attention to Gibbs measures defined on infinite trees. We will show that if LBP converges then the associated measures on the computation tree must converge to an element of $G(\gamma^\Phi)$.

**Limiting Gibbs Measures on Trees**

How do we construct an infinite volume Gibbs measure? So far we only have an implicit characterization via the DLR equations. For Gibbs measures defined on infinite trees we will show that the measures can arise as the weak limit of a sequence of Gibbs measures with fixed boundary conditions.

Let $\Phi$ be a pairwise potential whose Markov graph $(S^\Phi, E^\Phi)$ is a countably infinite tree. Let $T_n \subset S^\Phi$ be the set of nodes in the tree that are a distance of $n$ or less from the root. Recall $L_n \subset T_n$ is the set of nodes that are exactly a distance $n$ away from the root.

From the original potential $\Phi = \{\Phi_A\}$ we will now define a new sequence of potentials. For each $n \geq 1$ define $\Phi^{T_n} = \{\Phi_A^{T_n}\}$ by

$$\Phi_A^{T_n} = \begin{cases} \Phi_A, & \text{if } A \cap T_{n-1} \neq \emptyset; \\ \Phi_A^{bd,n} + \Phi_A, & \text{if } A \subset L_n, |A| = 1; \\ 0, & \text{otherwise.} \end{cases}$$

where $\Phi^{bd,n} \triangleq \{\Phi_A^{bd,n}\}$ represents added self-potentials at the leaves of the $T_n$ tree. If $\Phi_A^{bd,n} = 0$ for all $A \subset L_n$ then we call the boundary a *free boundary*. Note that $\gamma_{T_n}^{\Phi^{T_n}}(A \mid \omega)$ is independent of $\omega$ for $A \in \mathcal{F}_{T_n}$. For each $\Phi^{T_n}$ we can write the unique Gibbs measure as $\mu^{\Phi^{T_n}} = \lambda_{T_n^c} \gamma_{T_n}^{\Phi^{T_n}}$.

In the context of the computation tree $\mu^{\Phi^{T_n}}$ represents the measure on the infinite tree corresponding to $n$ iterations of LBP when initialized with the self-potentials $\Phi^{bd,n}$. We can relate the choice of $\Phi^{bd,n}$ to the choice of boundary self-potentials, $\{\Phi_{s_i}^{s_i \to s_j}\}_{(s_i, s_j) \in \vec{E}}$, in the LBP algorithm.

We are interested in conditions that insure the measures, $\{\mu^{\Phi^{T_n}}\}_{n \geq 1}$, converge to a limiting measure. The next proposition states that if they converge to a limiting measure then that measure must be an element of $G(\gamma^\Phi)$. Hence examining the structure of $G(\gamma^\Phi)$ is useful for determining convergence of LBP.

**Proposition 4.4** *Each subsequential limit of the sequence of measures $\{\mu^{\Phi^{T_n}}\}_{n \geq 1}$ belongs to $G(\gamma^\Phi)$.*

**Proof:** We sketch the proof. For full details see [3]. Note that for each $\Lambda \in \mathbb{S}$ there is an $n$ large enough such that $\gamma_\Lambda^{\Phi^{T_n}}(\cdot \mid \omega) = \gamma_\Lambda^{\Phi}(\cdot \mid \omega)$.

We need to show that the subsequential limits of the sequence of measures $\{\mu^{\Phi^{T_n}}\}_{n \geq 1}$ belong to $G(\Phi)$. Let $\mu$ be a subsequential limit where $\lim_{k \to \infty} \mu^{\Phi^{T_{n_k}}} = \mu$. We will show that $\mu$ satisfies the DLR equations. Let $\Lambda \in \mathbb{S}$ and let $A$ be a cylinder event. Then
$|\mu\gamma_\Lambda^\Phi(A) - \mu(A)|$

$$= \lim_{k \to \infty} |\lambda_{T_{n_k}^c} \gamma_{T_{n_k}}^{\Phi^{T_{n_k}}}(\gamma_\Lambda^\Phi(A)) - \lambda_{T_{n_k}^c} \gamma_{T_{n_k}}^{\Phi^{T_{n_k}}}(A)|$$
$$= \lim_{k \to \infty} |\lambda_{T_{n_k}^c} \gamma_{T_{n_k}}^{\Phi^{T_{n_k}}}(\gamma_\Lambda^\Phi(A)) - \lambda_{T_{n_k}^c} \gamma_{T_{n_k}}^{\Phi^{T_{n_k}}}(\gamma_\Lambda^{\Phi^{T_{n_k}}}(A))|$$
$$= 0$$

where the second equality holds for $k$ large enough so that $\Lambda \subset T_{n_k}$. □

**Markov Chains and Boundary Laws**

We have just shown that each subsequential limit of the sequence of measures $\{\mu^{\Phi^{T_n}}\}_{n \geq 1}$ belongs to $G(\gamma^\Phi)$. Here we describe the structure of the subsequential limits in terms of the Markov chains defined in $G(\gamma^\Phi)$.

We are given a potential $\Phi$ with a countably infinite tree Markov graph $(S^\Phi, E^\Phi)$. Let

$$S_{(i,j)}^\Phi \triangleq \{k \in S : d(k,j) = d(k,i) + 1\}$$

be the set of nodes in the "past" of the directed edge $(i, j)$ including the node $i$.

**Definition 4.2** *A measure $\mu^\Phi$ is a Markov chain on the tree if*

$$\mu(\omega_j \mid \mathcal{F}_{S_{(i,j)}^\Phi}) = \mu(\omega_j \mid \mathcal{F}_{\{i\}}) \quad \mu - a.s$$

*for all $(i,j) \in \vec{E}^\Phi$ and $\omega_j \in \mathcal{X}_j$.*

One can show that every Markov chain on the tree is a Markov field on the tree. The converse though does not always hold (see [1].) Measures that are Markov fields are often called *two-sided* whereas measures that are Markov chains are often called *one-sided*.

**Proposition 4.5** *Let $\Phi$ be a pairwise potential whose Markov graph is a tree. If $\mu$ is an extremal element of $G(\gamma^\Phi)$ then it is a Markov chain.*

**Proof:** See theorem 12.6 in [1]. □

There can exist Markov chains that are not extremal. Thus extremality alone is not enough to characterize the Markov chains in $G(\gamma^\Phi)$. It turns out, though, that we can characterize each Markov chain by the use of *boundary laws*. We will then show that these boundary laws are related to the messages in LBP.

First define for each $\{i,j\} \in E^\Phi$ a transfer matrix

$$Q_{\{i,j\}}(\omega_i, \omega_j) \triangleq e^{-\left(\Phi_{\{i,j\}}(\omega_i, \omega_j) + \frac{\Phi_{\{i\}}(\omega_i)}{|\partial i|} + \frac{\Phi_{\{j\}}(\omega_j)}{|\partial j|}\right)}$$

We can then write

$$\gamma_\Lambda(\omega_\Lambda \mid \omega) = Z_\Lambda(\omega)^{-1} \prod_{\{i,j\} \cap \Lambda \neq \emptyset} Q_{\{i,j\}}(\omega_i, \omega_j)$$

where here $Z_\Lambda(\omega) = \sum_{\omega_\Lambda} \prod_{\{i,j\} \cap \Lambda \neq \emptyset} Q_{\{i,j\}}(\omega_i, \omega_j)$.

**Definition 4.3** *A family $\{l_{(i,j)}\}_{(i,j) \in \vec{E}^\Phi}$, where each $l_{(i,j)}$ is a measure on $\mathcal{X}_i$ is called a* boundary law *if for each $(i,j) \in \vec{E}^\Phi$ we have $\forall \omega_i \in \mathcal{X}_i$:*

$$l_{(i,j)}(\omega_i) = \eta \prod_{k \in \partial i \setminus \{j\}} \sum_{\omega_k \in \mathcal{X}_k} Q_{\{k,i\}}(\omega_k, \omega_i) \, l_{(k,i)}(\omega_k) \quad (2)$$

Note the similarity to the message passing update rule.

**Proposition 4.6** *The following hold*

*(a) Each boundary law $\{l_{(i,j)}\}_{(i,j) \in \vec{E}^\Phi}$ for the transfer matrices $\{Q_{\{i,j\}}\}_{\{i,j\} \in E^\Phi}$ defines a unique Markov chain $\mu \in G(\gamma^\Phi)$ via the equation: for each connected set $\Lambda$*

$$\mu(\omega_{\Lambda \cup \partial \Lambda}) = \eta \prod_{k \in \partial \Lambda} l_{(k,k_\Lambda)}(\omega_k) \prod_{\{i,j\} \cap \Lambda \neq \emptyset} Q_{\{i,j\}}(\omega_i, \omega_j) \quad (3)$$

*(b) Each Markov chain $\mu \in G(\gamma^\Phi)$ admits a representation of the form (3) in terms of a boundary law $\{l_{(i,j)}\}_{(i,j) \in \vec{E}^\Phi}$ which is unique up to a positive scaling constant.*

**Proof:** See theorem 12.12 of [1]. □

**Boundary Laws and Messages**

Let $\Phi$ be a pairwise potential whose Markov graph is an infinite tree. We now relate boundary laws to messages. Recall that for each $(i,j) \in \vec{E}^\Phi$ we have

$$m_{(i,j)}(\omega_j) = \eta \sum_{\omega_i} e^{-(\Phi_{\{i,j\}}(\omega_i,\omega_j) + \omega_{\{i\}}(\omega_i))} \prod_{k \in \partial i \setminus j} m_{(k,j)}(\omega_i)$$

Note that equation (2) has a product of sums form whereas the equation above has a sum of products from. For each $(i,j) \in \vec{E}^\Phi$ let

$$\Phi_{\{i\}}^{i \to j}(\omega_i) = \eta \prod_{k \in \partial i \setminus j} m_{(k,i)}(\omega_i)$$

and after some algebra we get

$$\Phi_{\{i\}}^{i \to j}(\omega_i) = \eta \prod_{k \in \partial i \setminus j} \sum_{\omega_k} e^{-(\Phi_{\{k,i\}}(\omega_k,\omega_i) + \Phi_{\{k\}}(\omega_k))} \Phi_{\{k\}}^{k \to i}(\omega_k) \quad (4)$$

which is in a product of sums form. Comparing (4) and (2) we see that

$$l_{(i,j)}(\omega_i) = \eta e^{-(\frac{1}{|\partial i|}+1)} \Phi_{\{i\}}^{i \to j}(\omega_i)$$
$$= \eta e^{-\frac{1}{|\partial i|}-1} \prod_{k \in \partial i \setminus j} m_{(k,i)}(\omega_i)$$

Thus we can go from boundary self-potentials and messages to boundary laws and vice-versa.

**Proposition 4.7** *Each subsequential limit of the sequence of measures $\{\mu^{\Phi^{T_{n_k}}}\}_{k \geq 1}$ corresponds to a Markov chain in $G(\gamma^\Phi)$.*

**Proof:** We sketch the proof. See [3] for full details. Let $\{n_k\}_{k \geq 1}$ be a subsequence for which $\{\mu^{\Phi^{T_{n_k}}}\}_{k \geq 1}$ converges to some measure $\mu$. By proposition 4.4 this $\mu$ is an element of $G(\gamma^\Phi)$. By proposition 3.1 the messages $\{m_{(i,j)}^{n_k}\}$ converge along the subsequence $\{n_k\}$ to some fixed point solution $\{m_{(i,j)}^*\}$. By equation (2) and (4) this fixed point solution corresponds to a boundary law. By proposition 4.6 we see that $\mu$ is a Markov chain. □

Recall that proposition 4.2 states that the set $G(\gamma^\Phi)$ is nonempty and hence contains at least one extremal element. By proposition 4.5 this extremal element is a Markov chain. These results along with propositions 3.1 and 4.7 give us a new way to show that there always exists at least one solution to the LBP fixed point equations.

It has been observed in practice that LBP sometimes oscillates. In this case LBP is actually jumping between different solutions of the LBP fixed point equations and hence jumping between different Markov chains defined on the computation tree.

In summary, we have characterized each subsequential limit measure corresponding to the LBP algorithm in terms of a Markov chain defined on the computation tree. We will now discuss conditions that insure the existence of a unique limit.

## 5 Unique Gibbs Measure Case

Here we consider the case of a unique Gibbs measure: $|G(\gamma^\Phi)| = 1$. Clearly there can be only be one subsequential limit of the sequence of measures $\mu^{\Phi T_n}$. Hence LBP converges. We can say something stronger though: LBP converges uniformly over the choice of all initializing messages.

**Proposition 5.1** *If $|G(\gamma^\Phi)| = 1$ then*

$$\lim_{n \to \infty} \gamma_{T_n}^\Phi(\cdot \mid \omega) = \mu^\Phi(\cdot) \text{ uniformly in } \omega \in \Omega.$$

**Proof:** See proposition 7.11 in [1]. □

**Proposition 5.2** *If $|G(\gamma^\Phi)| = 1$ then for any cylinder event $A$ we have*

$$\lim_{n \to \infty} \mu_{T_n}^{\Phi T_n}(A) = \mu^\Phi(A)$$

*uniformly over the boundary self-potentials.*

**Proof:** We sketch the proof. See [3] for full details. Let $\{\Phi^{bd,n}\}_{n \geq 1}$ be any set of boundary self-potentials. Let $A$ be $\mathcal{F}_{\Lambda_1}$-measurable and choose $\epsilon > 0$. Then by proposition 5.1 there exists a $\Lambda_2 \supset \Lambda_1$ such that $|\gamma_{\Lambda_2}^\Phi(A \mid \omega) - \mu^\Phi(A)| \leq \epsilon$. Now for $T_n \supset \Lambda_2$ we have

$$\begin{aligned}
|\mu_{T_n}^{\Phi T_n}(A) - \mu^\Phi(A)| &= |\mu_{T_n}^{\Phi T_n} \gamma_{\Lambda_2}^{\Phi T_n}(A \mid \omega) - \mu^\Phi(A)| \\
&= |\mu_{T_n}^{\Phi T_n} \gamma_{\Lambda_2}^\Phi(A \mid \omega) - \mu^\Phi(A)| \\
&= |\mu_{T_n}^{\Phi T_n} (\gamma_{\Lambda_2}^\Phi(A \mid \omega) - \mu^\Phi(A))| \\
&\leq |\mu_{T_n}^{\Phi T_n} \epsilon| \\
&= \epsilon
\end{aligned}$$

where the first equality holds via the DLR equation. Thus the convergence rate is independent of the boundary self-potentials used. □

In summary if $|G(\Phi)| = 1$ then LBP will converge uniformly over the boundary self-potentials. Next we present Dobrushin's sufficient condition for uniqueness of the limiting Gibbs measure.

**Dobrushin's Condition**

A Gibbs potential can lead to many different phases if nodes that are far apart from each other do not mix fast enough. Dobrushin proposed the following condition that insures fast mixing and hence uniqueness.

**Proposition 5.3** *Let $\Phi$ be a pairwise potential. If*

$$\sup_{i \in S} \sum_{A \ni i} (|A| - 1)\delta(\Phi_A) < 2 \quad (5)$$

*then $|G(\gamma^\Phi)| = 1$. Where $\delta(f) \triangleq \sup_x f(x) - \inf_x f(x)$.*

**Proof:** See proposition 8.8 in [1]. □

This proposition states that the "influence" node $i$ has on the rest of the nodes depends on two things: the number of neighbors it has and the strength of the potentials, measured by $\delta(\Phi)$, it takes part in. Note that the self-potentials do not play a part in Dobrushin's condition.

Let us return to the issue of LBP. Let $\Phi$ be the potential for the finite Gibbs measure that we wish to apply LBP to. Let $\tilde{\Phi}$ be the corresponding potential on the computation tree. The local topology of the computation tree looks like the local topology of the original graph. Hence to show $|G(\tilde{\Phi})| = 1$ we need to show

$$\sup_{\tilde{i} \in S^{\tilde{\Phi}}} \sum_{A \ni \tilde{i}} (|A| - 1)\delta(\tilde{\Phi}_A) = \max_{i \in S^\Phi} \sum_{A \ni i} (|A| - 1)\delta(\Phi_A) < 2$$

Note that the maximum condition is very easy to check on finite graphs.

**Rate of Convergence**

Here we give a condition on the rate of convergence of LBP. Let $c(\Phi) = \max_{s \in S^\Phi} \sum_{A \ni s} e^2(|A| - 1)\delta(\Phi_A)$.

**Proposition 5.4** *If $c(\Phi) < 2$ then*

$$|\mu_{T_n}^{\Phi^{T_n}}(\omega_s) - \mu^\Phi(\omega_s)| \leq k e^{-n}$$

*where $s$ is the root node and $k = (1 - c(\Phi))^{-1}$ is a constant.*

**Proof:** See theorem 8.23, remark 8.26, and corollary 8.32 of [1]. □

## 6 Conclusions

In this paper we have introduced tools from the theory of Gibbs measures to analyze the convergence properties of the LBP algorithm. In particular we related the problem of convergence of LBP to the existence of a weak limit for a sequence of Gibbs measures defined the corresponding computation tree. We have introduced a condition that insures the uniqueness of the Gibbs measure defined on the infinite computation tree. Hence this condition insures LBP converges.

**Acknowledgments**

The authors would like to thank Sanjoy Mitter, Kevin Murphy, and Mark Paskin for many helpful discussion.

This work was supported by the DoD Multidisciplinary University Research Initiative (MURI) program administered by the Office of Naval Research under Grant N00014-00-1-0637.## References

[1] H. O. Georgii, *Gibbs Measures and Phase Transitions*. Berlin, Walter de Gruyter and Co., 1988.

[2] F. Jensen, *An Introduction to Bayesian Networks*. UCL Press, London, 1996.

[3] S. Tatikonda and M. Jordan, " Conditions for Convergence in the Loopy Belief Propagation Algorithm." Berkeley working paper, 2002.

[4] M. Wainwright, T. Jaakkola and A. Willsky, "Tree-Based Reparameterization for Approximate Estimation on Loopy graphs." Advances in Neural Information Processing Systems 14, 2002.

[5] Y. Weiss, "Correctness of Local Probability Propagation in Graphical Models with Loops." *Neural Computation*, 12:1-41, 2000.

[6] J. Yedidia, W. Freeman. Y. Weiss, "Bethe Free Energy, Kikuchi Approximations and Belief Propagation Algorithms." Advances in Neural Information Processing Systems 13, 2000.

[7] A. Yuille, "A Double-Loop Algorithm to Minimize the Bethe and Kikuchi Free Energies." *Neural Computation*, to appear, 2001.